\documentclass[sn-mathphys-num]{sn-jnl}

\usepackage{graphicx}%
\usepackage{multirow}%
\usepackage{amsmath,amssymb,amsfonts}%
\usepackage{amsthm}%
\usepackage{mathrsfs}%
\usepackage[title]{appendix}%
\usepackage{xcolor}%
\usepackage{textcomp}%
\usepackage{manyfoot}%
\usepackage{booktabs}%
\usepackage{algorithm}%
\usepackage{algorithmicx}%
\usepackage{algpseudocode}%
\usepackage{listings}%
\usepackage{multirow}%
\usepackage{float}
\usepackage{geometry}
\usepackage{booktabs}
\usepackage{pdfpages}

\theoremstyle{thmstyleone}%
\theoremstyle{thmstyletwo}%

\theoremstyle{thmstylethree}%

\begin{document}


\title{Rethinking Plant Disease Diagnosis: Bridging the Academic–Practical Gap with Vision Transformers and Zero-Shot Learning}

\author[1]{Wassim Benabbas}
\author[2,3]{Mohammed Brahimi}
\author[4,5]{Samir Akhrouf}
\author[6,7]{Bilal Fortas}

\affil[1]{Department of Computer Science, Mohamed El Bachir El Ibrahimi University, Bordj Bou Arreridj, Algeria.}
\affil[2]{Intelligent Systems Enginnering, National School of Artificial Intelligence (ENSIA), Algiers, Algeria.}
\affil[3]{Laboratory of Physical Chemistry and Biological Materials, Skikda, Algeria.}
\affil[4]{Laboratory of Informatics and its Applications of M'sila, Mohamed Boudiaf University, M'Sila, Algeria.}
\affil[5]{Department of Computer Science, Mohamed Boudiaf University, M'Sila, Algeria}
\affil[6]{Department of agricultural science,  Mohamed El Bachir El Ibrahimi University, Bordj Bou Arreridj, Algeria.}
\affil[7]{Laboratoire d’Amélioration et de Développement de la Production Végétale et Animale (LADPVA), University of Ferhat ABBAS (UFAS-Sétif1), Sétif, Algeria.}


\abstract {Recent advances in deep learning have enabled significant progress in plant disease classification using leaf images. Much of the existing research in this field has relied on the PlantVillage dataset, which consists of well-centered plant images captured against uniform, uncluttered backgrounds. Although models trained on this dataset achieve high accuracy, they often fail to generalize to real-world field images, such as those submitted by farmers to plant diagnostic systems. This has created a significant gap between published studies and practical application requirements, highlighting the necessity of investigating and addressing this issue. In this study, we investigate whether attention-based architectures and zero-shot learning approaches can bridge the gap between curated academic datasets and real-world agricultural conditions in plant disease classification. We evaluate three model categories: Convolutional Neural Networks (CNNs), Vision Transformers, and Contrastive Language–Image Pre-training (CLIP)–based zero-shot models. While CNNs exhibit limited robustness under domain shift, Vision Transformers demonstrate stronger generalization by capturing global contextual features. Most notably, CLIP models classify diseases directly from natural language descriptions without any task-specific training, offering strong adaptability and interpretability. These findings highlight the potential of zero-shot learning as a practical and scalable domain adaptation strategy for plant health diagnosis in diverse field environments.}




\maketitle

\section{Introduction}\label{sec1}
Global food security is under threat as plant diseases cause up to 40\% of crop losses annually \cite{ref1}. Crops such as potatoes and tomatoes are vital for both nutrition and the economy, yet they remain vulnerable to devastating diseases like early blight and late blight\cite{ref2}. These diseases not only reduce crop yields but also contribute to environmental degradation through excessive pesticide use. As a result, safeguarding plant health is critical for promoting sustainable agriculture and ensuring global food security \cite{ref18}.



Early detection of plant diseases is key to minimizing damage and ensuring food supply resilience \cite{ref17}. However, manual diagnosis in agricultural fields is time-consuming, labor-intensive, and often requires expert knowledge, which may not be available in rural or resource-limited settings \cite{ref15} . To overcome these limitations, researchers have increasingly turned to artificial intelligence (AI) and deep learning to automate the diagnosis and classification of plant diseases from images \cite{ref24}.

CNNs have been widely used for this task due to their ability to extract spatial features from leaf images. For instance, Mohanty et al. \cite{ref3} demonstrated high classification accuracy using AlexNet and GoogleNet on the PlantVillage dataset. Too et al. \cite{ref4} explored multiple CNN architectures, including VGG16, ResNet50, and InceptionV3, showing strong performance in plant disease classification. Similarly, Brahimi et al. \cite{ref5} applied deep learning to tomato leaf diseases using transfer learning, achieving robust results across multiple disease classes.

More recently, transformer-based models have emerged as powerful alternatives to CNNs. Models such as the Vision Transformer (ViT-B/16 and ViT-B/32) have been applied successfully to agricultural tasks. Dosovitskiy et al. \cite{ref6} introduced ViT for image classification, and studies like Hassani et al. \cite{ref7} and Khan et al. \cite{ref8} have applied ViT variants to plant disease datasets with promising accuracy and generalization. Additionally, Swin Transformers have also been explored for fine-grained disease classification due to their hierarchical design.

Beyond traditional visual models, vision-language approaches like CLIP leverage both text and images, enabling zero-shot classification by aligning disease descriptions with visual features. Radford et al. \cite{ref9} first introduced CLIP, and recent studies such as Yu et al. \cite{ref10}, and Zhang et al. \cite{ref11} explored its potential in agriculture, demonstrating that combining semantic information with visual cues can enhance performance on unseen or real-world disease samples.


While deep learning has achieved good results in plant disease classification, most existing studies rely on datasets collected in highly controlled environments, such as uniform lighting, isolated leaves, and plain backgrounds \cite{ref25}. A widely used example is the PlantVillage dataset \cite{ref12}, which contains clean images with minimal background noise. However, models trained under these ideal conditions often exhibit substantial performance degradation when applied to real-world images captured by farmers in natural settings, where background complexity, lighting variation, and multiple overlapping objects introduce significant noise \cite{ref13}\cite{ref14}. This study is motivated by our previous work \cite{ref16}, which demonstrated that convolutional neural networks (CNNs) trained exclusively on academic datasets often exhibit limited generalizability to real-world scenarios. In that work, we examined the effectiveness of realistic data augmentation strategies in improving CNN training. Although the proposed approach outperformed conventional augmentation techniques, it remained insufficient in substantially enhancing CNN performance under complex, real-world conditions.

This study addresses a key challenge in plant disease classification: the performance gap between models trained on curated academic datasets and their generalization to complex, real-world agricultural conditions. We hypothesize that attention-based mechanisms in transformer models and the use of zero-shot learning via CLIP can significantly enhance robustness and adaptability to such scenarios.

To test these hypotheses, we explore three categories of models: CNN-based, Transformer-based, and CLIP-based zero-shot architectures. All supervised models are trained using the widely used PlantVillage dataset, while their performance is assessed on a diverse, field-collected test set that reflects the variability of real-world farming environments. In contrast, CLIP-based models are not fine-tuned on any plant-specific data. Instead, they classify images using descriptive text prompts, offering key advantages in terms of flexibility, interpretability, and ease of integrating previously unseen disease classes.

By validating the potential of zero-shot learning in reducing data dependency and improving real-world applicability, our study aims to propose scalable, practical solutions for plant health monitoring beyond traditional benchmark evaluations.


\section{Materials and Methods}\label{sec2}
\subsection{Methodology}
\begin{figure}[ht]
    \centering
    \includegraphics[
        page=1, 
        trim=0.1in 0.1in 0.1in 0.1in, 
        clip,
        width=\textwidth
    ]{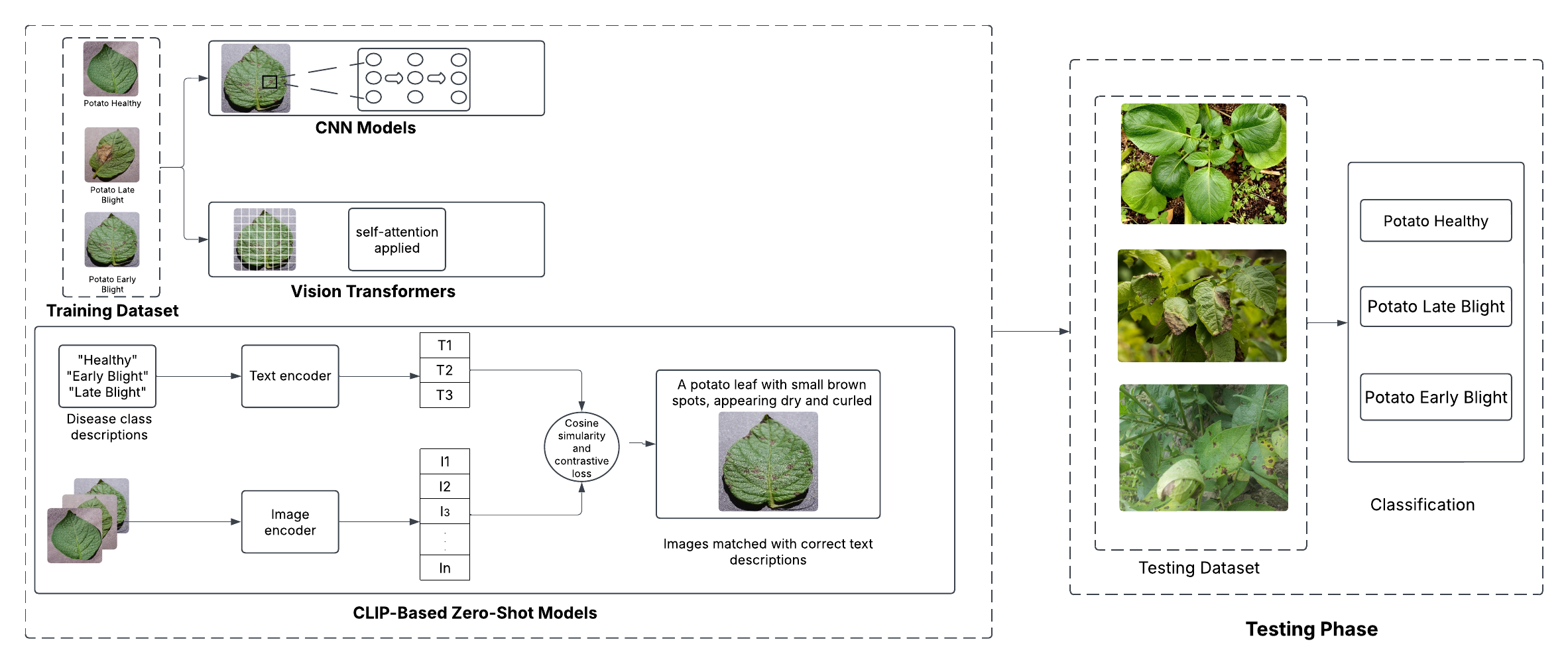}
    \caption{Model Categorization and Evaluation Pipeline for Plant Disease Classification}
    \label{method}
\end{figure}

To validate the aforementioned hypothesis, we adopt a comprehensive methodology that leverages three distinct model categories: CNN-based architectures, Transformer-based networks, and CLIP-based zero-shot models, as illustrated in Fig.~\ref{method}.

\subsubsection{CNN-Based Models}
We include three state-of-the-art convolutional architectures: EfficientNet-B0, ResNet50, and InceptionV3. These models follow the classical deep learning pipeline, where input images are processed through stacked convolutional layers to extract hierarchical spatial features. These layers exploit local receptive fields, weight sharing, and translation invariance—hallmarks of CNNs in visual tasks.

CNN-based approaches remain dominant in plant disease classification research due to their strong performance on clean, labeled datasets such as PlantVillage \cite{ref3,ref13}. However, their reliance on local features may limit their generalization in field conditions where disease patterns are more variable and occluded.

\subsubsection{Transformer-Based Models}
Transformer architectures have emerged as a powerful alternative to CNNs by modeling images as sequences of non-overlapping patches. Each patch is embedded and enriched with positional encodings before being processed by multi-head self-attention mechanisms. In our work, we adopt ViT-B/16, ViT-B/32, Swin-T, and Swin-S architectures.

We hypothesize that the attention mechanism enables these models to selectively attend to symptomatic regions while ignoring irrelevant visual noise, which is particularly beneficial in complex field scenarios. This global context modeling may offer superior robustness over CNNs for real-world leaf imagery. This experiment investigates whether Transformer-based architectures can effectively capture long-range dependencies and highlight disease-specific symptoms across diverse plant classes. Recent studies have demonstrated the potential of Transformers for fine-grained classification in agricultural contexts and other vision tasks \cite{ref22,ref6}.


\subsubsection{CLIP-Based Zero-Shot Models} 
To explore generalization beyond traditional supervised learning, we incorporate CLIP-based zero-shot models (CLIP-ViT-B/16 and CLIP-ViT-B/32) into our study. These models align textual disease descriptions with image embeddings in a shared semantic space via contrastive pretraining. At inference, handcrafted prompts (e.g., “A healthy potato leaf with no spots”, “A potato leaf with concentric brown lesions”, “A potato leaf with irregular dark patches”) are encoded by the text encoder, while plant images are processed by the vision encoder. The resulting embeddings are compared through cosine similarity, and the highest similarity score determines the predicted class. 

\begin{figure}[ht]
    \centering
    \includegraphics[
        page=1, 
        trim=0.1in 0.1in 0.1in 0.1in, 
        clip,
        width=\textwidth
    ]{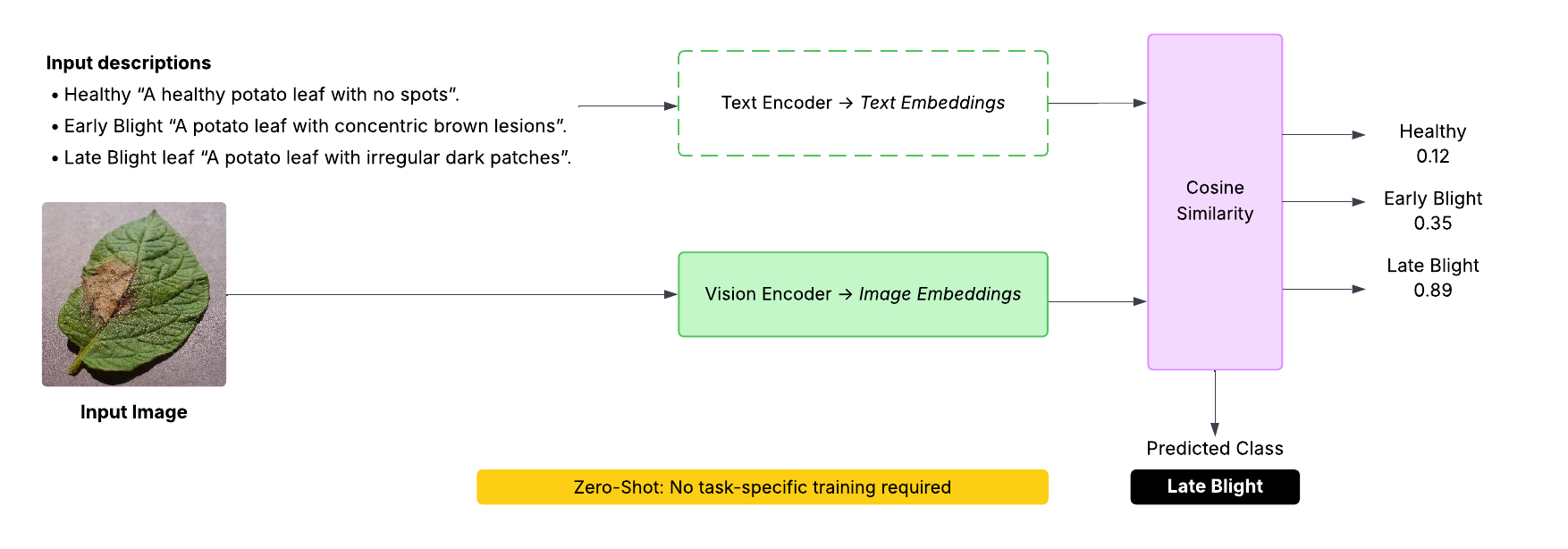}
    \caption{Model Categorization and Evaluation Pipeline for Plant Disease Classification}
    \label{Clip_zero_shot}
\end{figure}

As illustrated in Figure~\ref{Clip_zero_shot} and Algorithm~\ref{alg:clip_sum}, the CLIP zero-shot pipeline operates by comparing input images against multiple textual descriptions without requiring any task-specific fine-tuning. In this workflow, disease-related prompts are first encoded by the text encoder and stored as fixed embeddings, while each new image is encoded online by the vision encoder. The embeddings are then projected into a shared semantic space, where cosine similarity determines the most likely class. This organization highlights an important optimization: the text encoder is executed only once, whereas the variable component across inferences is the image embedding.

To further enhance robustness, multiple textual variants can be defined for each class, and their similarity scores aggregated, as shown in Algorithm~\ref{alg:clip_sum}. This strategy mitigates sensitivity to prompt wording and improves classification stability. For example, in the illustrative case of Figure~\ref{Clip_zero_shot}, a potato leaf affected by late blight achieves similarity scores of (Healthy: 0.12, Early Blight: 0.35, Late Blight: 0.89), leading to the correct prediction of Late Blight. These values are provided solely to demonstrate how CLIP performs zero-shot classification in practice.

Overall, this paradigm eliminates the need for large annotated datasets, making it especially valuable in agricultural contexts where expert labeling is limited or costly. While CLIP has shown broad generalization in vision-language domains \cite{ref9,ref23}, applying it to plant disease classification remains largely unexplored, representing a novel contribution of this work.


\begin{algorithm}[H]
\caption{Zero-Shot CLIP Classification with Multiple Descriptions per Class}
\label{alg:clip_sum}
\begin{algorithmic}[1]
\Require Image $I$, set of textual descriptions $\{d_c^j\}_{j=1}^{N_c}$ for each class $c = 1,\dots,C$
\Ensure Predicted label $\hat{y}$

\State \textbf{(1) Text Encoding (offline, one-time):}
\For{each class $c = 1,\dots,C$}
    \For{each description $d_c^j, \; j = 1,\dots,N_c$}
        \State $t_c^j \gets f_{\text{text}}(d_c^j)$
        \State $\tilde{t}_c^j \gets \dfrac{t_c^j}{\|t_c^j\|}$
    \EndFor
\EndFor

\State \textbf{(2) Image Encoding (online, per input):}
\State $v \gets f_{\text{img}}(I)$
\State $\tilde{v} \gets \dfrac{v}{\|v\|}$

\State \textbf{(3) Similarity Computation:}
\For{each class $c = 1,\dots,C$}
    \State $S_c \gets \sum_{j=1}^{N_c} \tilde{v}^\top \tilde{t}_c^j$
\EndFor

\State \textbf{(4) Classification:}
\State $\hat{y} \gets \arg\max_{c} S_c$

\State \Return $\hat{y}$

\end{algorithmic}
\end{algorithm}

\subsubsection{Training Strategy and Evaluation Setup}

The CNN and Transformer models are trained in a supervised manner using a dataset taken from PlantVillage, which includes labeled potato leaf images across three classes: Early Blight, Late Blight, and Healthy. Each model is trained following its respective architecture and optimization procedure.

By contrast, CLIP-based zero-shot models (CLIP-ViT-B/16 and CLIP-ViT-B/32) are not fine-tuned on this dataset. Instead, they use a pretrained vision-language model that encodes both images and descriptive disease prompts into a shared embedding space, learned through contrastive training on large-scale image-text data. At inference time, the model selects the class whose text embedding is most similar to that of the input image.

This setup enables us to assess how well each model performs in complex, practical agricultural scenarios—whether by generalizing from curated training data (CNNs and Transformers) or through zero-shot inference without task-specific training (CLIP).



\subsection{Data Structure}
This study utilizes two distinct datasets: one for model training and another for evaluation.  

\subsubsection{Training Dataset} \label {train_dataset0}
\begin{figure}[ht]
    \centering
    \includegraphics[
        page=1, 
        trim=0.1in 0.1in 0.1in 0.1in, 
        clip,
        width=\textwidth
    ]{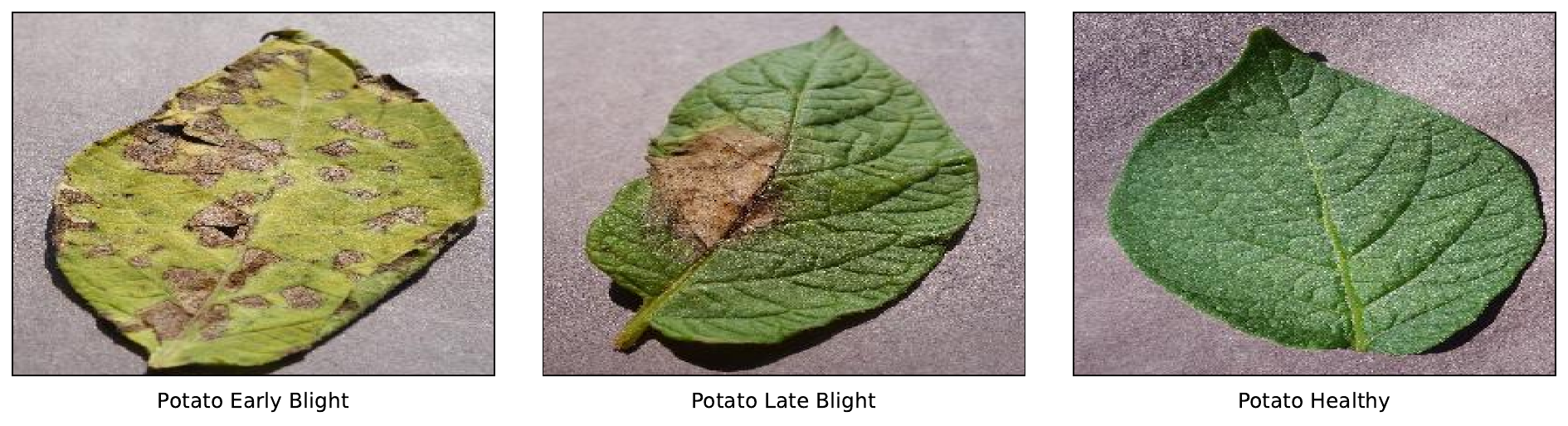}
    \caption{Sample images from the Training Dataset}
    \label{Train_img}
\end{figure}

The training data is derived from the PlantVillage collection and specifically targets potato leaf conditions. It comprises three categories: Healthy, Early Blight, and Late Blight. Table \ref{Table1} summarizes the dataset composition, while Fig. \ref{Train_img} displays representative examples.

\begin{table}[ht]
  \centering
  \caption{Distribution of classes in the Training Dataset}
  \label{Table1}
  \begin{tabular}{lc}
    \toprule
    \textbf{Disease Class} & \textbf{Number of Images} \\
    \midrule
    Healthy      & 152   \\
    Early Blight & 1000  \\
    Late Blight  & 1000  \\
    \midrule
    \textbf{Total} & \textbf{2152} \\
    \bottomrule
  \end{tabular}
\end{table}


\subsubsection{Testing Dataset}
The testing dataset consists of 945 real-world images collected from four different sources. These images are used to evaluate the performance of the models in realistic and challenging conditions. The dataset includes images with complex backgrounds, varying lighting conditions, and diverse camera qualities, reflecting the real-world scenarios faced by farmers. Table \ref{Table2} presents an overview of the testing dataset.

\begin{table}[h]
  \centering
  \caption{Testing Data Overview}
  \label{Table2}
  \scriptsize
  \renewcommand{\arraystretch}{0.95} 
  \setlength{\tabcolsep}{5pt} 
  \begin{tabular}{llc}
    \toprule
    \textbf{Source Dataset} & \textbf{Class Label} & \textbf{Image Count} \\
    \midrule
    \multirow{3}{*}{Farmy} & Early Blight & 34 \\
                           & Late Blight & 58 \\
                           & Healthy     & 132 \\
                           \cmidrule(lr){2-3}
                           & \textbf{Total} & \textbf{224} \\
    \midrule
    \multirow{2}{*}{Africa} & Late Blight & 68 \\
                            & Healthy     & 26 \\
                            \cmidrule(lr){2-3}
                            & \textbf{Total} & \textbf{94} \\
    \midrule
    \multirow{3}{*}{Peru} & Early Blight & 71 \\
                          & Late Blight  & 254 \\
                          & Healthy      & 21 \\
                          \cmidrule(lr){2-3}
                          & \textbf{Total} & \textbf{346} \\
    \midrule
    \multirow{3}{*}{Internet} & Early Blight & 98 \\
                              & Late Blight  & 100 \\
                              & Healthy      & 83 \\
                              \cmidrule(lr){2-3}
                              & \textbf{Total} & \textbf{281} \\
    \midrule
    \multirow{3}{*}{\textbf{Overall}} & Early Blight & 203 \\
                                      & Late Blight  & 480 \\
                                      & Healthy      & 262 \\
                                      \cmidrule(lr){2-3}
                                      & \textbf{Grand Total} & \textbf{945} \\
    \bottomrule
  \end{tabular}
\end{table}

\begin{enumerate}[1.]

\item \textbf{Farmy Dataset:} This dataset was collected via the Farmy mobile application, which leverages AI-based methods to detect plant diseases instantly from leaf photos. It contains 224 labeled images divided into three categories: Early Blight, Late Blight, and Healthy. To ensure data reliability, all images were reviewed and confirmed by a plant disease specialist. A sample of these images is shown in Fig.~\ref{Farmy_img}.

\begin{figure}[ht]
    \centering
    \includegraphics[
        page=1, 
        trim=0.1in 0.1in 0.1in 0.1in, 
        clip,
        width=\textwidth
    ]{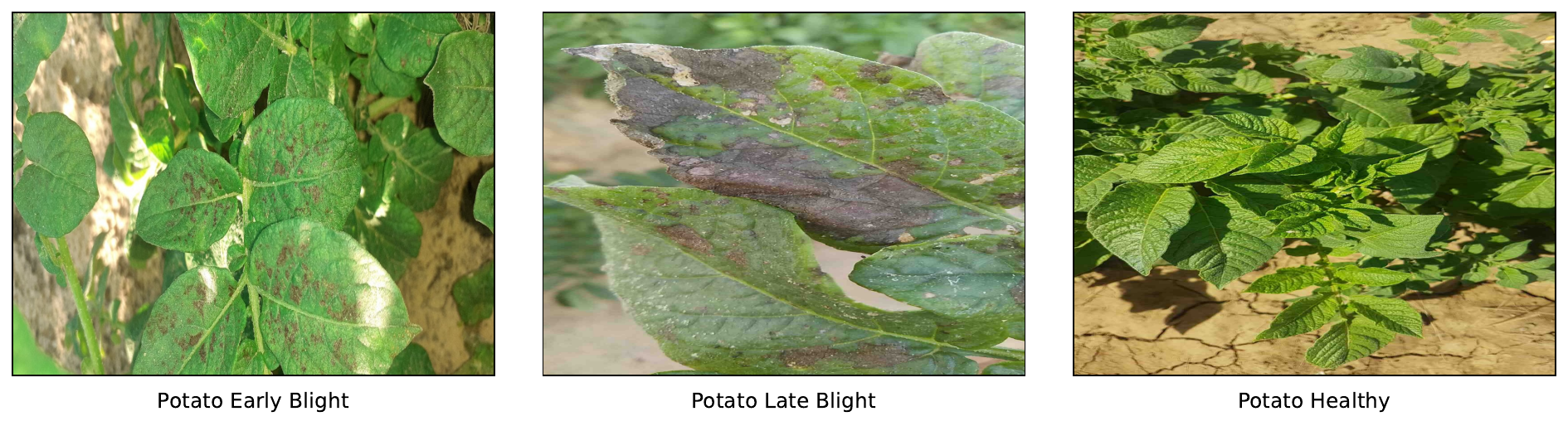}
    \caption{Sample images from the Farmy Dataset}
    \label{Farmy_img}
\end{figure}

%



\item \textbf{Africa Dataset:} Introduced by \cite{ref19}, this dataset contains 94 images captured in African regions. It covers two categories: Healthy and Late Blight. Representative samples are illustrated in Fig.~\ref{Africa_img}.

\begin{figure}[ht]
    \centering
    \includegraphics[
        page=1, 
        trim=0.1in 0.1in 0.1in 0.1in, 
        clip,
        width=\textwidth
    ]{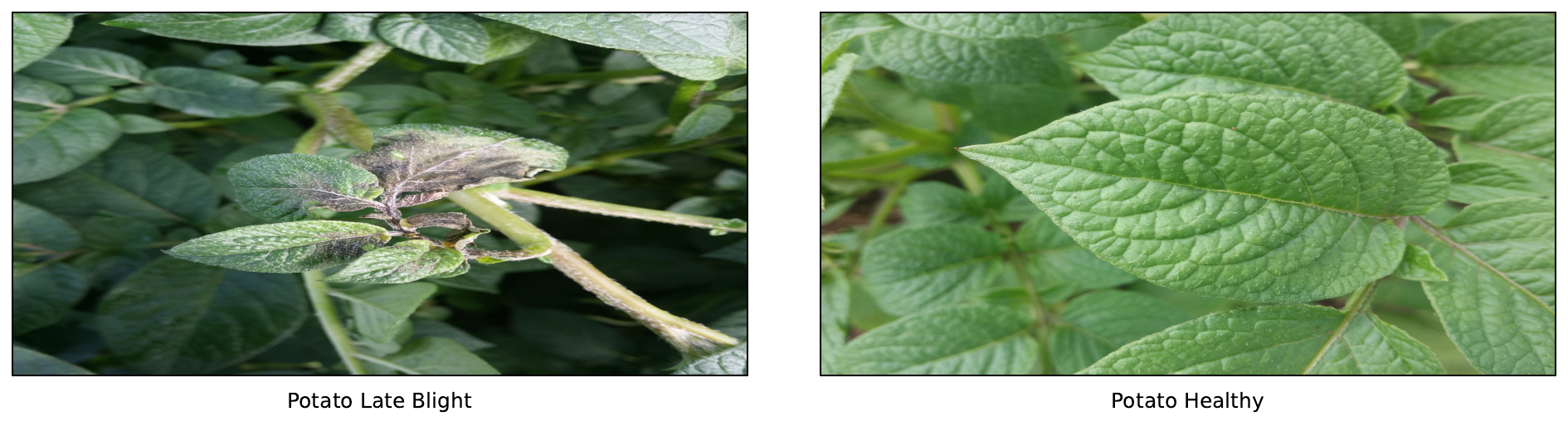}
    \caption{Sample images from the Africa Dataset}
    \label{Africa_img}
\end{figure}

%


\item \textbf{Peru Dataset:} also presented by \cite{ref19}, contains 346 images taken in Peru. It includes three classes: Early Blight, Late Blight, and Healthy. Fig. \ref{Peru_img} displays a sample of the extracted images.

\begin{figure}[ht]
    \centering
    \includegraphics[
        page=1, 
        trim=0.1in 0.1in 0.1in 0.1in, 
        clip,
        width=\textwidth
    ]{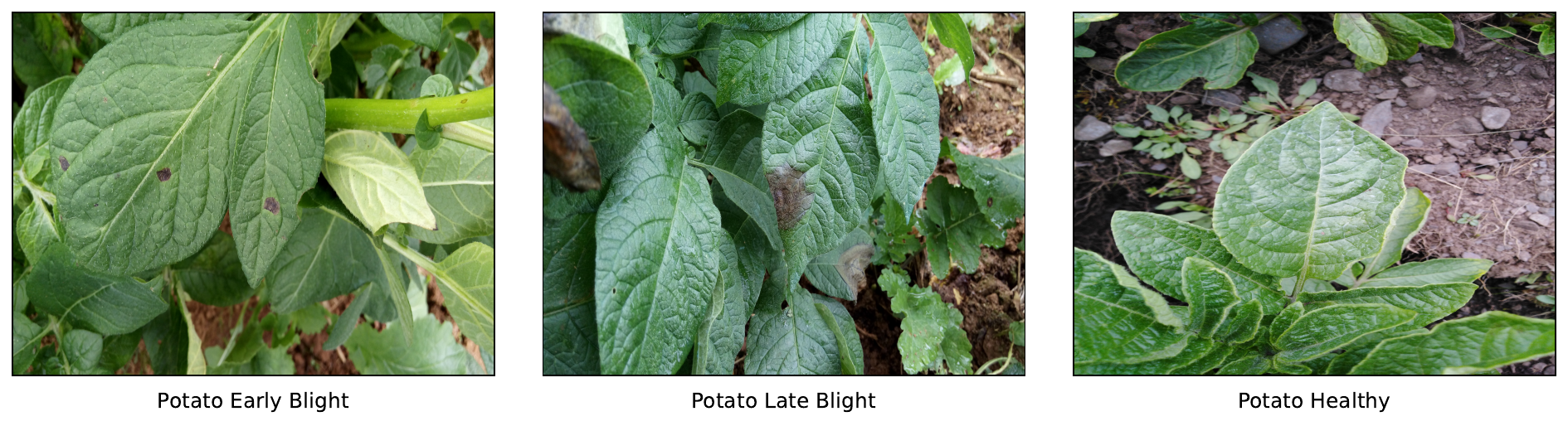}
    \caption{Sample images from the Peru Dataset}
    \label{Peru_img}
\end{figure}

%


\item \textbf{Internet Dataset:} was collected from publicly available sources and validated by an expert in plant leaf diseases. It contains 281 images across three classes: Early Blight, Late Blight, and Healthy. Fig. \ref{Net_img} showcases sample images from this dataset.

\begin{figure}[ht]
    \centering
    \includegraphics[
        page=1, 
        trim=0.1in 0.1in 0.1in 0.1in, 
        clip,
        width=\textwidth
    ]{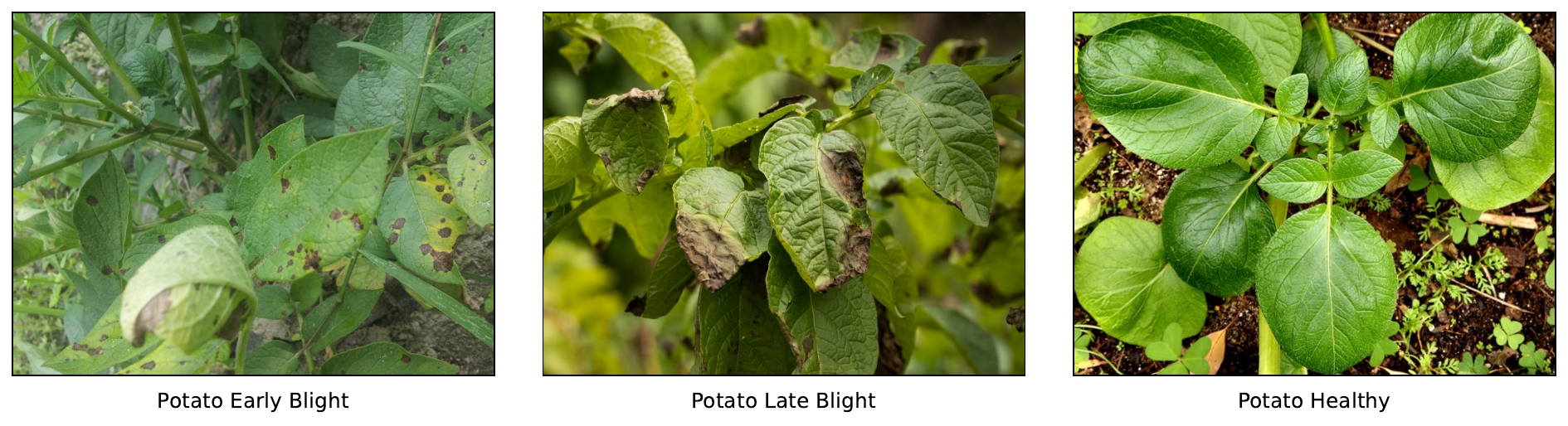}
    \caption{Sample images from the Internet Dataset}
    \label{Net_img}
\end{figure}

%


\subsection{Models}
We evaluate nine models categorized into three groups: CNNs, Vision Transformers, and Zero-Shot CLIP models.

\begin{enumerate}
    \item \textbf{CNN Models:}
    \begin{enumerate}[1.]
        \item \textbf{EfficientNet-B0} – A compound-scaled CNN with 5.3M parameters.
        \item \textbf{ResNet50} – A 50-layer deep residual network with 25.5M parameters.
        \item \textbf{InceptionV3} – A CNN utilizing multi-scale inception modules, containing 23.8M parameters.
    \end{enumerate}

    \item \textbf{Vision Transformers:}
    \begin{enumerate}[1.]
        \item \textbf{ViT-B-16} – A 12-layer transformer processing 16×16 image patches, with 86M parameters.
        \item \textbf{ViT-B-32} – A variant using 32×32 patches, maintaining 86M parameters.
        \item \textbf{Swin-S} – A shifted window transformer architecture with 50M parameters.
        \item \textbf{Swin-T} – A lightweight Swin Transformer variant with 28M parameters.
    \end{enumerate}

    \item \textbf{Zero-Shot CLIP Models:}
    \begin{enumerate}[1.]
        \item \textbf{CLIP-ViT-B-16} – A zero-shot model that combines a ViT-B-16 image encoder with a Transformer-based text encoder.
        \item \textbf{CLIP-ViT-B-32} –A zero-shot model that combines a ViT-B-32 image encoder with a Transformer-based text encoder.
    \end{enumerate}
\end{enumerate}
\end{enumerate}
These CLIP models are not fine-tuned for the specific task. Instead, they perform zero-shot classification by calculating the cosine similarity between image features and the textual representations of each class. The following prompts were embedded for each class to enable effective visual-semantic alignment:

\begin{itemize}
    \item \textbf{Potato Early Blight:}
    \begin{itemize}
        \item This is a photo of a potato leaf with concentric brown spots surrounded by yellow halos, typical of early blight.
        \item This is a potato plant with dry, curling lower leaves covered in small dark lesions caused by early blight.
        \item This image shows early blight symptoms: brown circular lesions on aging potato foliage.
        \item This potato leaf has target-like spots with yellowish borders, starting from the bottom of the plant.
        \item A potato plant with early blight infection showing dry, brittle leaves and small, scattered necrotic spots.
        \item Potato foliage exhibiting dark brown specks and yellowing, characteristic of fungal early blight.
    \end{itemize}
    
    \item \textbf{Potato Late Blight:}
    \begin{itemize}
        \item This is a photo of a potato leaf with large, irregular dark lesions and water-soaked edges caused by late blight.
        \item Late blight on potato leaves appears as black, greasy patches that spread rapidly in wet conditions.
        \item This is a potato plant with collapsing leaves and fuzzy white mold on the underside, typical of late blight.
        \item Potato foliage with late blight shows irregular brown patches, often starting at leaf tips or edges.
        \item This potato leaf displays soft, dark lesions and a wilted appearance under humid conditions.
        \item An advanced late blight infection causes leaf tissue to rot, with visible fungal growth and necrosis.
    \end{itemize}
    
    \item \textbf{Potato Healthy:}
    \begin{itemize}
        \item This is a photo of a healthy potato leaf, fully green with no visible disease or damage.
        \item A vibrant, intact potato plant with fresh green leaves and no signs of infection.
        \item This potato leaf is smooth, evenly colored, and free from any brown or yellow spots.
        \item Healthy potato foliage with firm structure, no curling, and consistent green pigmentation.
        \item This is an image of a disease-free potato plant under natural light, showing perfect leaf condition.
        \item A top-down view of a healthy potato plant with all leaves intact and bright green.
    \end{itemize}
\end{itemize}

The models used in this study are summarized in Table \ref{tab:model_specs}.
\begin{table}[h]
\centering
\caption{Model Architecture Summary}
\label{tab:model_specs}
\begin{tabular}{lll}
\toprule
\textbf{Category} & \textbf{Model} & \textbf{Parameters} \\
\midrule
CNN & EfficientNet-B0 & 5.3M \\
& ResNet50 & 25.5M \\
& InceptionV3 & 23.8M \\
\midrule
Transformer & ViT-B-16 & 86M \\
& ViT-B-32 & 86M \\
& Swin-S & 50M \\
& Swin-T & 28M \\
\midrule
Zero-shot (CLIP) & CLIP-ViT-B-16 & 149M (86M image + 63M text) \\
& CLIP-ViT-B-32 & 149M (86M image + 63M text) \\
\bottomrule
\end{tabular}
\end{table}

\subsection{Training and Fine-Tuning}

\subsubsection{Pre-training}
Represents the training of deep learning models on the massive ImageNet dataset \cite{ref20}, which includes over 14 million images across 1,000 object categories. These pre-trained models capture universal visual features that serve as excellent starting points for specialized tasks. By leveraging this large-scale pre-training, we benefit from:
\begin{itemize}
    \item Learned hierarchical feature representations
    \item Robust low-level edge and texture detectors
    \item High-level semantic understanding of visual patterns
\end{itemize}

\subsubsection{Fine-tuning}

Fine-tuning adapts pre-trained models to our specific plant disease classification task by replacing the final classification layer with a new one tailored to our three target classes: Early Blight, Late Blight, and Healthy. This process:

\begin{itemize}
    \item Retains the general feature extraction capabilities learned from ImageNet
    \item Specializes the model for agricultural image analysis
    \item Reduces the amount of training data required compared to training from scratch
\end{itemize}

To achieve optimal adaptation, all models except Zero-Shot CLIP were fine-tuned on the Training Dataset described in Sect. \ref{train_dataset0}, which consists of PlantVillage images spanning the three disease classes. By using this dataset, we aim to bridge the gap between general visual features and domain-specific disease patterns, ensuring the models effectively recognize and differentiate plant health conditions.

\subsubsection{Optimization}
We employed the Adam optimizer \cite{ref21}, which combines the advantages of adaptive learning rates with momentum-based updates. The optimizer automatically adjusts:
\begin{itemize}
    \item Learning rates for individual parameters
    \item Momentum terms for gradient descent stability
    \item Update magnitudes based on historical gradients
\end{itemize}
The Adam optimizer updates the model parameters \(\theta\) as follows:
\[
\theta \leftarrow \theta - \eta \frac{m_t}{\sqrt{v_t} + \epsilon}
\]
In this equation, \(\eta\) denotes the learning rate, \(\epsilon\) is a small constant added for numerical stability, \(m_t\) represents the estimate of the first moment (mean), and \(v_t\) corresponds to the second moment estimate (uncentered variance).

\subsubsection{Loss Function}
The categorical cross-entropy loss measures the discrepancy between predicted class probabilities and true disease labels. This function:
\begin{itemize}
    \item Penalizes incorrect classifications proportionally to their confidence
    \item Encourages the model to output well-calibrated probabilities
    \item Works effectively with the Adam optimizer's adaptive updates
\end{itemize}
In multi-class classification tasks, the categorical cross-entropy loss \(\mathcal{L}_{\text{CCE}}\) is expressed as:
\[
\mathcal{L}_{\text{CCE}}(\hat{y}, y) = -\sum_{c=1}^{C} y_c \log(\hat{y}_c)
\]
with \(C\) denoting the number of classes, \(y_c\) representing the ground truth label, and \(\hat{y}_c\) the predicted probability for class \(c\).

\subsection{Evaluation}

We implemented 5-fold cross-validation to obtain statistically reliable performance estimates, with all metrics averaged across folds. The following evaluation measures were computed:

\begin{itemize}
    \item \textbf{Macro Precision} - Evaluates the precision for each disease class individually and computes their average. A higher value indicates fewer false positives (e.g., healthy leaves incorrectly labeled as diseased):
    \[
    P_{\text{macro}} = \frac{1}{3}\left(\frac{TP_E}{TP_E+FP_E} + \frac{TP_L}{TP_L+FP_L} + \frac{TP_H}{TP_H+FP_H}\right)
    \]
    Where $E$=Early Blight, $L$=Late Blight, $H$=Healthy.

    \item \textbf{Macro Recall} - Measures the model's ability to detect all cases of each disease. High recall means fewer missed infections (diseased leaves misclassified as healthy):
    \[
    R_{\text{macro}} = \frac{1}{3}\left(\frac{TP_E}{TP_E+FN_E} + \frac{TP_L}{TP_L+FN_L} + \frac{TP_H}{TP_H+FN_H}\right)
    \]

    \item \textbf{Macro F1-score} - Balances precision and recall into a single metric. Critical for evaluating both disease detection accuracy and avoidance of false alarms:
    \[
    F1_{\text{macro}} = \frac{1}{3}\left(2\frac{P_E R_E}{P_E+R_E} + 2\frac{P_L R_L}{P_L+R_L} + 2\frac{P_H R_H}{P_H+R_H}\right)
    \]

    \item \textbf{Confusion Matrix} - Shows exact classification patterns, revealing which diseases are most frequently confused (e.g., Early vs Late Blight):
    \[
    \begin{bmatrix}
    TP_E & FP_{E\to L} & FP_{E\to H} \\
    FP_{L\to E} & TP_L & FP_{L\to H} \\
    FP_{H\to E} & FP_{H\to L} & TP_H
    \end{bmatrix}
    \]

    \item \textbf{Receiver Operating Characteristic / Area Under the Curve (ROC/AUC)} - Evaluates the model's ability to rank diseased leaves higher than healthy leaves across all possible decision thresholds. Higher AUC means better disease prioritization:
    \begin{itemize}
        \item Plots TPR (True Positive Rate) vs FPR (False Positive Rate)
        \item Macro-AUC = $\frac{AUC_E + AUC_L + AUC_H}{3}$
    \end{itemize}

    \item \textbf{Matthews Correlation Coefficient (MCC)} - Provides a single balanced measure (-1 to 1) accounting for all correct and incorrect classifications. Ideal for imbalanced disease datasets:
    \[
    MCC = \frac{\sum_{i=E,L,H}(TP_i TN_i - FP_i FN_i)}{\sqrt{\prod_{j=E,L,H}(TP_j+FP_j)(TP_j+FN_j)(TN_j+FP_j)(TN_j+FN_j)}}
    \]
\end{itemize}




\section{Results} \label{sec3}
We present the performance of various deep learning models trained on curated PlantVillage images and tested on complex real-world field conditions. 

 Table~\ref{tab:model_performance} presents the performance of all evaluated models based on macro F1-score, macro recall, and macro precision. Among all models, the CLIP-ViT-B-16 model achieved the best overall performance with a macro F1-score of 66.29\%, demonstrating strong generalization in a zero-shot setting by effectively leveraging semantic descriptions of disease classes. The CLIP-ViT-B-32 model also performed well, achieving an F1-score of 56.63\%, outperforming several fine-tuned vision transformers and CNNs.
 
 Among the transformer-based models, ViT-B-16 followed closely with an F1-score of 61.41\%, while Swin-S achieved 52.70\%. CNN architectures showed comparatively lower performance, with EfficientNet-B0 being the best in this group at 33.71\%. ResNet50 exhibited the weakest performance overall, with an F1-score of 30.15\%.
 
 These results highlight clear distinctions in model capabilities for disease classification under real-world conditions.

\begin{table}[h]
\centering
\caption{Performance of Deep Learning Models Based on Macro F1-Score, Macro Recall, and Macro Precision}
\label{tab:model_performance}
\begin{tabular}{lcccc}
\toprule
Group & Model & Macro Precision & Macro Recall & Macro F1-score \\
\midrule
\multirow{3}{*}{CNN Models} 
& EfficientNet-B0 & 41.00 & 45.17 & 33.71 \\
& InceptionV3 & 57.32 & 42.37 & 32.25 \\
& ResNet50 & 56.80 & 40.80 & 30.15 \\
\midrule
\multirow{4}{*}{Vision Transformers} 
& ViT-B-16 & 75.43 & 61.02 & 61.41 \\
& ViT-B-32 & 61.39 & 48.84 & 47.06 \\
& Swin-S & 63.14 & 55.55 & 52.70 \\
& Swin-T & 63.14 & 44.75 & 35.19 \\
\midrule
\multirow{2}{*}{Multimodal CLIP Models} 
& CLIP-ViT-B-16 & 67.30 & 66.21 & 66.29 \\
& CLIP-ViT-B-32 & 64.38 & 57.94 & 56.63 \\
\bottomrule
\end{tabular}
\end{table}

\begin{figure}[ht]
    \centering
    \includegraphics[
        page=1, 
        trim=0.1in 0.1in 0.1in 0.1in, 
        clip,
        width=\textwidth
    ]{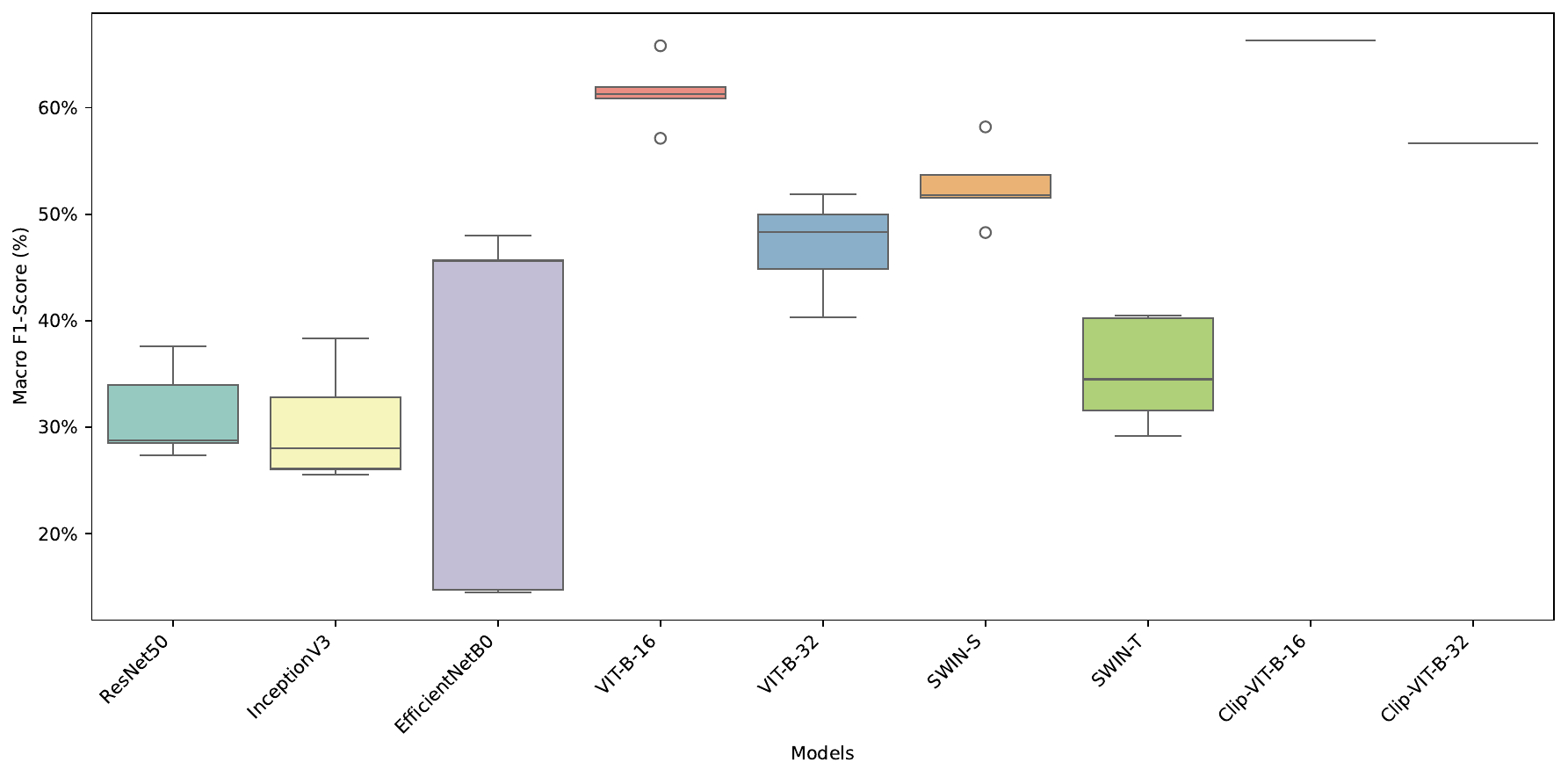}
    \caption{F1-Score Distribution Across 5 Folds for Model Performance Comparison}
    \label{f1_score_boxplot_real_data}
\end{figure}

The boxplot visualization in Fig. \ref{f1_score_boxplot_real_data} illustrates the variation in macro f1-score across five folds for each model, highlighting differences in performance stability and generalization. Since the CLIP models operate in a zero-shot setting, their results are not based on cross-validation, and thus only a single macro F1-score is reported for each. Among the fine-tuned models, ViT-B-16 achieved the strongest overall consistency, with F1-scores ranging from 57.13\% to 65.82\%, indicating both high performance and moderate variability across folds. ViT-B-32 displayed wider variation (40.29\% to 51.86\%), suggesting less stable generalization. Swin-S maintained moderate performance with tighter distribution, while Swin-T showed the highest variability among transformers, with scores spanning 29.15\% to 40.48\%. CNN-based models exhibited the greatest instability. EfficientNet-B0, the best-performing CNN, had F1-scores ranging from 14.47\% to 48.03\%, reflecting considerable inconsistency. ResNet50 fluctuated between 27.37\% and 37.63\%, and InceptionV3 dropped as low as 25.54\%, highlighting weaker generalization in real-world conditions.

%

\begin{figure}[ht]
    \centering
    \includegraphics[
        page=1, 
        trim=0.1in 0.1in 0.1in 0.1in, 
        clip,
        width=\textwidth
    ]{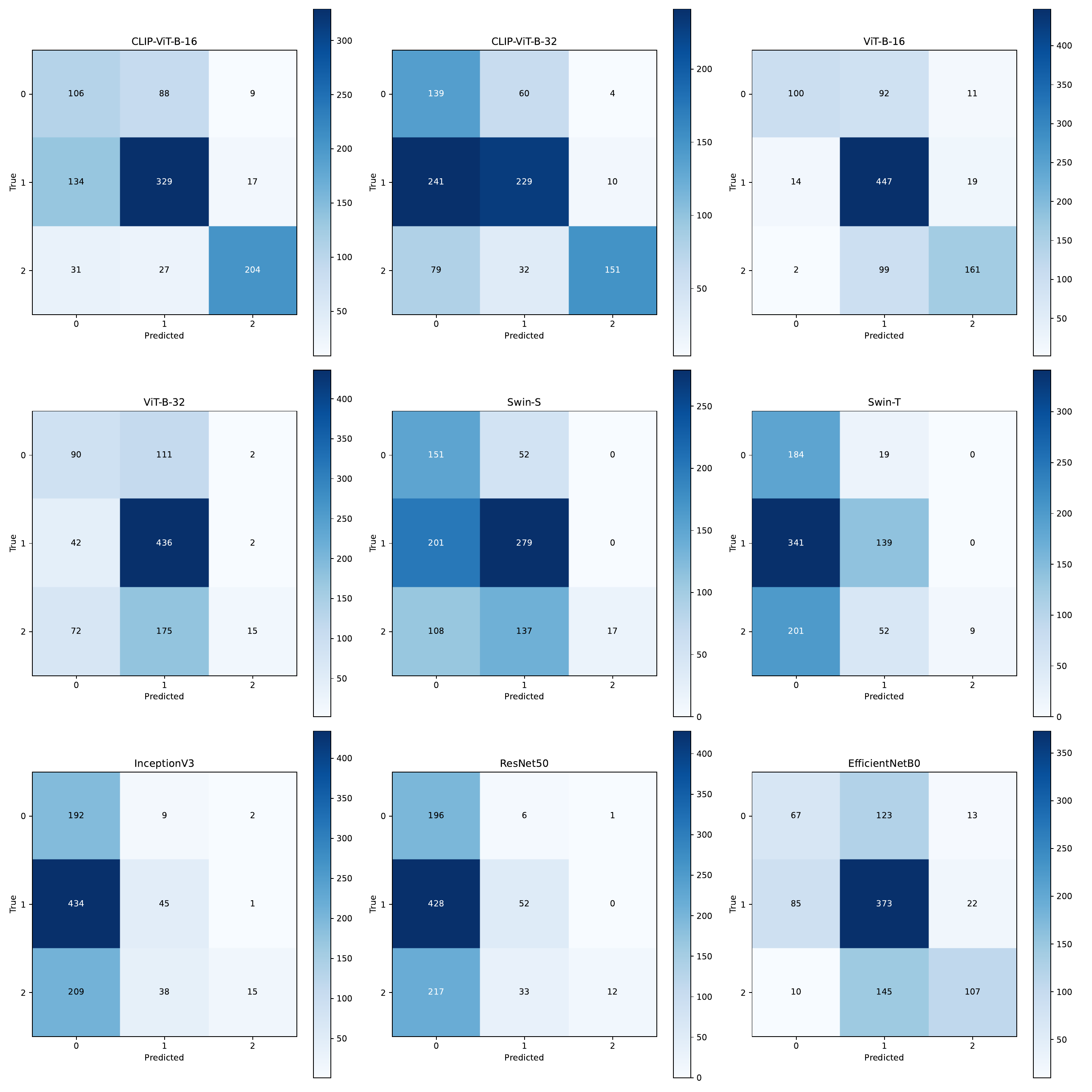}
    \caption{Comparison of Confusion Matrices Across Models}
    \label{conf_matrices}
\end{figure}

The confusion matrices presented in Fig. \ref{conf_matrices} reveal distinct classification patterns across models, with class indices defined as follows: 0 for Potato Early Blight, 1 for Potato Late Blight, and 2 for Potato Healthy. CLIP-ViT-B-16 achieved the most balanced performance, correctly identifying most Potato Late Blight samples while showing minor misclassification between Potato Early Blight and Potato Late Blight. ViT-B-16 followed closely, demonstrating strong classification for Potato Late Blight but with some confusion between Potato Early Blight and Potato Healthy samples. Among transformer models, Swin-S showed moderate accuracy, while Swin-T struggled, frequently misclassifying Potato Late Blight as Potato Early Blight. The CNN models, particularly ResNet50 and InceptionV3, exhibited the highest misclassification rates, with many Potato Late Blight samples incorrectly labeled as Potato Early Blight, indicating difficulty in distinguishing disease classes. However, EfficientNet-B0 performed relatively better among CNNs, showing improved classification accuracy while still struggling with class separation compared to transformers.

\begin{figure}[ht]
    \centering
    \includegraphics[
        page=1, 
        trim=0.1in 0.1in 0.1in 0.1in, 
        clip,
        width=\textwidth
    ]{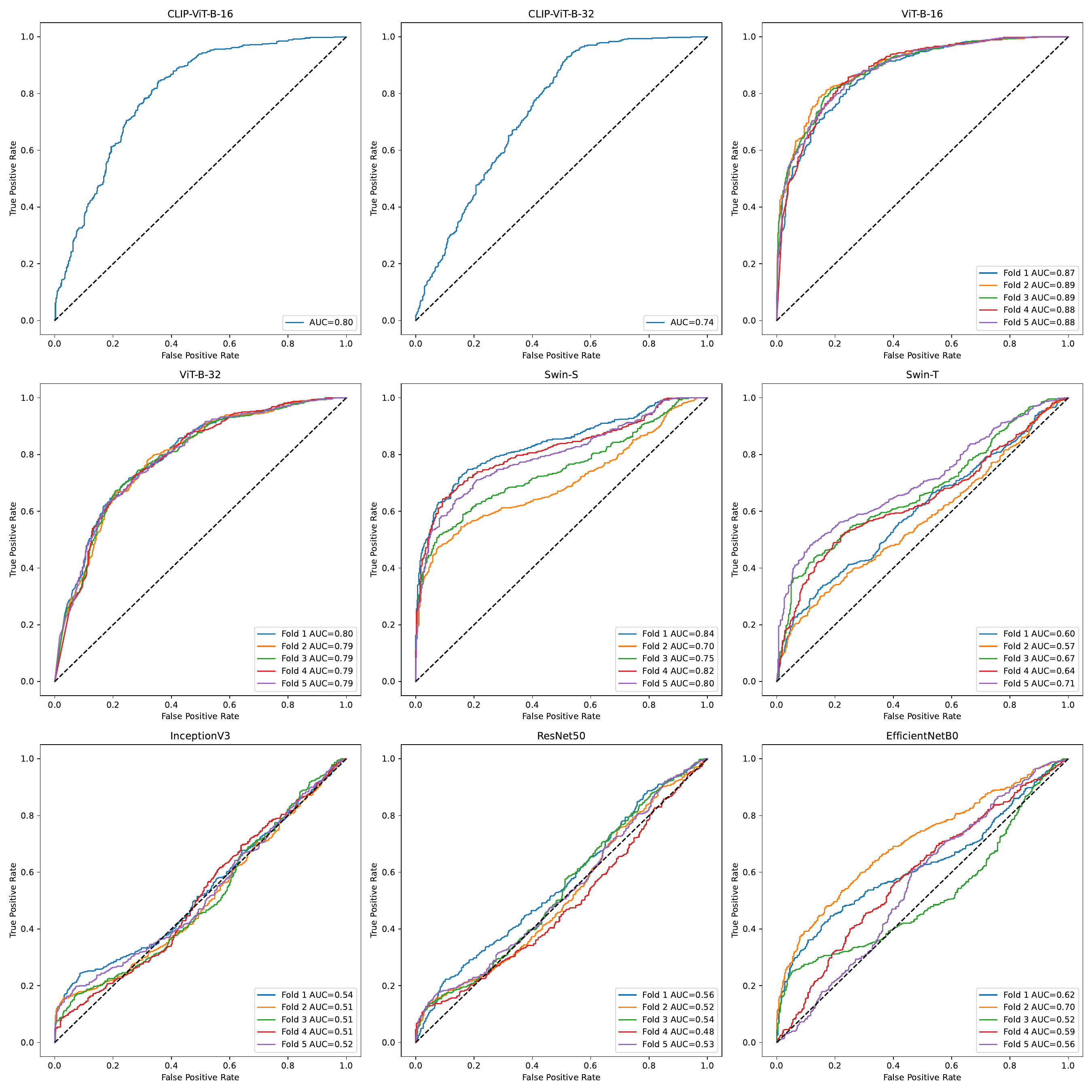}
    \caption{ROC Curves and AUC Scores for Model Comparison}
    \label{roc_curves}
\end{figure}

In Fig.~\ref{roc_curves}, the ROC curves and AUC values reveal clear distinctions in model performance across architectures. ViT-B-16 achieved the highest AUC values (0.87–0.89) across all folds, indicating strong overall classification ability. In contrast, the CLIP models—evaluated under a zero-shot setting—were tested using only a single fold. Despite this, CLIP-ViT-B-16 attained a strong AUC of 0.80, and CLIP-ViT-B-32 followed with an AUC of 0.74. Among vision transformers, ViT-B-32 maintained stable AUC values (0.79–0.80), but slightly lower than its ViT-B-16 counterpart. Swin-S showed inconsistent performance, with AUC values varying between 0.70 and 0.84, while Swin-T had weaker performance, fluctuating between 0.57 and 0.71, indicating lower reliability in classification. CNN models exhibited the lowest AUC values. EfficientNet-B0 performed best among CNNs, with AUC values ranging from 0.52 to 0.70, but still lagged behind transformers. ResNet50 (AUC: 0.48–0.56) and InceptionV3 (AUC: 0.51–0.54) struggled the most, reflecting their limitations in distinguishing between disease classes.

%

\begin{figure}[ht]
    \centering
    \includegraphics[
        page=1, 
        trim=0.1in 0.1in 0.1in 0.1in, 
        clip,
        width=\textwidth
    ]{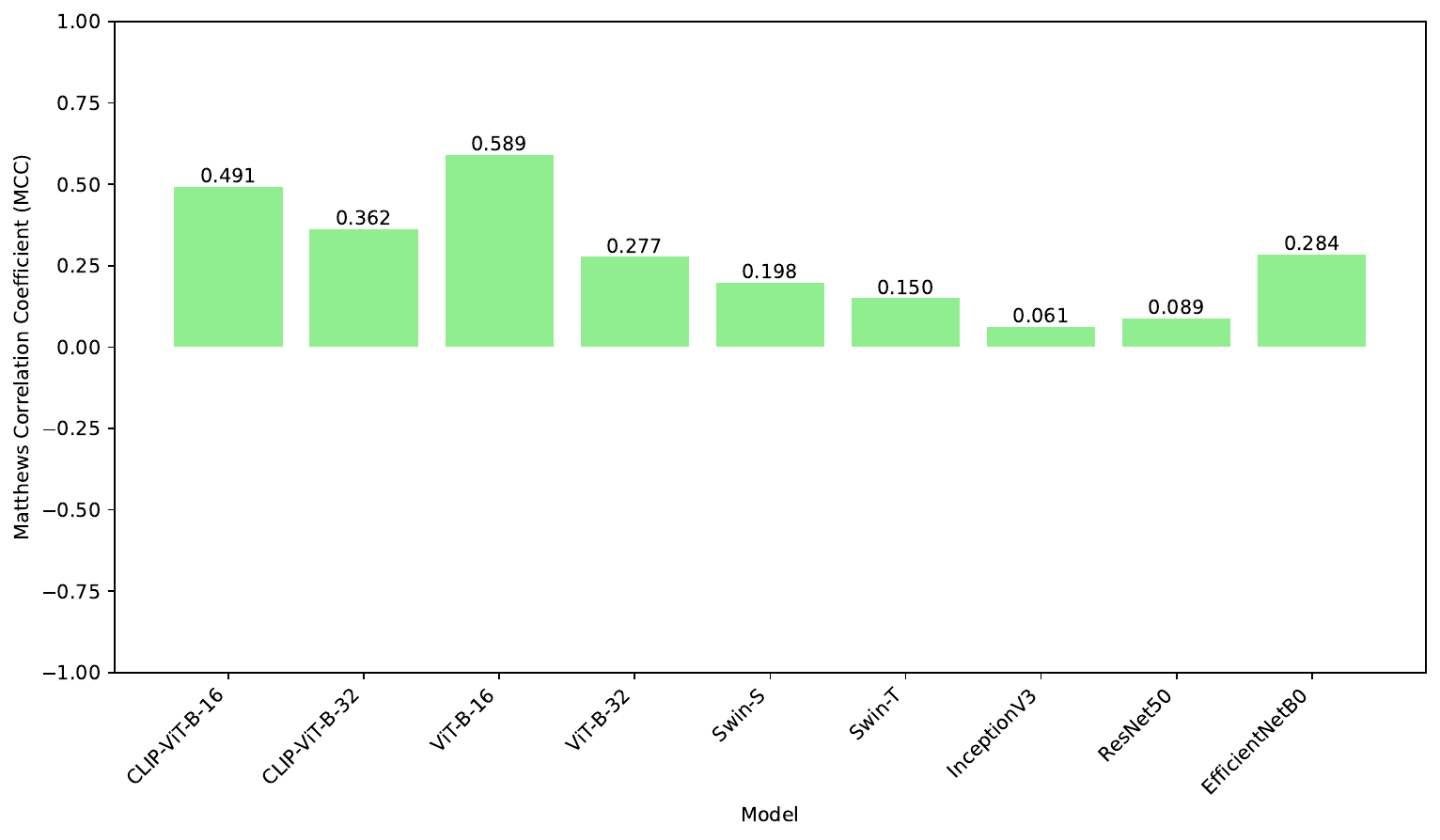}
    \caption{Comparison of MCC Scores Across Models}
    \label{fig:mcc_scores}
\end{figure}

As shown in Fig.~\ref{fig:mcc_scores}, the MCC scores provide further insights into model reliability in plant disease classification. ViT-B-16 achieved the highest MCC score (0.589), reflecting its strong correlation between predicted and true labels. CLIP-ViT-B-16 followed closely with an MCC of 0.491, underscoring its balanced and effective performance, even in a zero-shot setting. ViT-B-32 achieved a moderate MCC of 0.277, while CLIP-ViT-B-32 outperformed it slightly with a score of 0.362, highlighting the potential of multimodal information in improving prediction alignment. Among the Swin architectures, Swin-S scored 0.198, outperforming Swin-T, which attained an MCC of 0.150. CNN-based models recorded the weakest correlations overall. EfficientNet-B0 led this group with an MCC of 0.284, whereas InceptionV3 (0.061) and ResNet50 (0.089) demonstrated very limited predictive consistency, reinforcing the superior reliability of transformer and zero-shot models in complex disease classification tasks.

\begin{figure}[ht]
    \centering
    \includegraphics[
        page=1, 
        trim=0.1in 0.1in 0.1in 0.1in, 
        clip,
        width=\textwidth
    ]{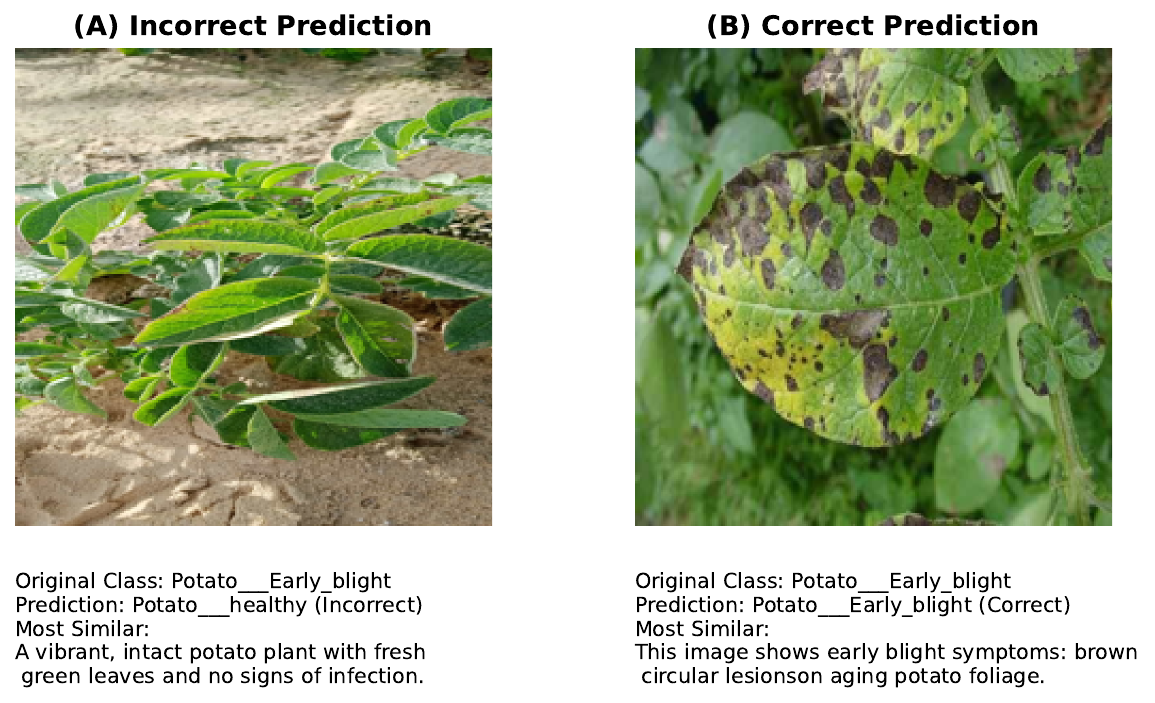}
    \caption{CLIP Interpretability}
    \label{clip_Interpretability}
\end{figure}

%
To further evaluate the interpretability of CLIP in the context of potato disease detection, we conducted a case study using images belonging to the Early blight class. Two contrasting cases are illustrated in Figure \ref{clip_Interpretability}.

In the first case, the image was misclassified as “Potato healthy”, with the most similar textual description being “A vibrant, intact potato plant with fresh green leaves and no signs of infection.” Although the ground truth label was Early blight, the captured image did not clearly display the characteristic symptoms, which led to a confusion with the healthy class.

In the second case, the image was correctly classified as “Potato Early blight”, with the most similar textual description being “This image shows early blight symptoms: brown circular lesions on aging potato foliage.” The textual description accurately matched the visual symptoms present in the image.

These two examples illustrate how CLIP interpretability provides both a classification output and an associated natural language description.

\section{Discussion}

In this section, we analyze the strengths and limitations of different deep learning architectures in plant disease classification, emphasizing their generalization capabilities when trained on controlled datasets and evaluated on complex, real-world images, and from zero-shot learning. The results highlight the superiority of transformer-based and CLIP-based zero-shot models over traditional CNNs, with key insights drawn from performance metrics such as macro F1-score, AUC, MCC, and confusion matrices.

\subsection{Comparative Performance of CNNs, Transformers, and CLIP Zero-Shot Models}

Table~\ref{tab:model_performance} highlights a significant performance gap between CNNs and transformer-based models, particularly under real-world conditions. CNN architectures such as EfficientNet-B0 (33.71\% F1-score), InceptionV3 (32.25\% F1-score), and ResNet50 (30.15\% F1-score) showed limited generalization when faced with noisy backgrounds, varied lighting, and natural leaf distortions. Their reliance on localized spatial features makes them effective in controlled environments but less robust in diverse field scenarios.

In contrast, transformer-based models demonstrated much stronger generalization. ViT-B-16 achieved a 61.41\% F1-score, benefiting from its self-attention mechanism that captures global context and long-range dependencies. The CLIP-based models, especially CLIP-ViT-B-16 (66.29\% F1-score), surpassed all others. This strong performance is due to their ability to align images with descriptive class texts in a shared embedding space, enabling effective zero-shot classification.

CLIP-based zero-shot models were the top performers, with CLIP-ViT-B/16 achieving an F1-score of 66.29\%. Unlike CNNs and Transformers, CLIP does not rely on retraining or labeled task-specific data. Instead, it leverages its pretrained vision-language embedding space to compare image features with textual descriptions. This enables classification using only semantic similarity, without the need for annotated examples from the target domain. The strong results achieved by CLIP models highlight the potential of zero-shot learning as a practical solution for scalable plant disease diagnosis—especially in scenarios where labeled data is limited, costly, or unavailable.

\subsection{Stability and Generalization Across Folds}

The boxplot visualization in Fig.~\ref{f1_score_boxplot_real_data}  further highlights the stability of different architectures across five folds. Transformer-based models such as ViT-B-16 consistently demonstrated strong generalization capabilities, with moderate variability and high classification scores, reflecting their ability to extract context-aware features through self-attention mechanisms.
In contrast, CNN models displayed greater fluctuations. EfficientNet-B0 ranged from 14.47\% to 48.03\%, reflecting inconsistent learning patterns. ResNet50 and InceptionV3 showed even higher instability, struggling to maintain robust performance across different data partitions.
CLIP models, evaluated in a zero-shot setting, were tested using only a single fold. Their performance depended heavily on the quality and precision of the textual descriptions provided, demonstrating the flexibility and adaptability of the zero-shot approach when paired with well-crafted semantic prompts.
These results highlight that transformers and CLIP models are not only more accurate but also more consistent, thanks to their deeper contextual understanding and, in the case of CLIP, the capacity to generalize from descriptive language inputs.

\subsection{Classification Errors and Model Robustness}

The confusion matrices in Fig.~\ref{conf_matrices} reveal distinct classification patterns across architectures. CLIP-ViT-B-16 and ViT-B-16 demonstrated superior classification ability, with fewer misclassifications between Potato Early Blight and Potato Late Blight. However, Swin-T and CNN models exhibited significant misclassification errors, frequently confusing Potato Late Blight with Potato Early Blight, indicating a lack of discriminative power.

This challenge arises from CNNs' difficulty in capturing subtle textural variations in diseased leaves, whereas transformers leverage global feature dependencies, making them more effective at distinguishing disease types. The advantage of CLIP models lies in their ability to integrate textual disease descriptions, further refining classification accuracy.

\subsection{Model Discriminability: Insights from AUC and MCC Scores}

The ROC curves and AUC values in Fig.~\ref{roc_curves} reinforce the findings from macro F1-score analysis. ViT-B-16 achieved the highest AUC values (0.87–0.89), followed by CLIP-ViT-B-16 (0.80), demonstrating their superior ability to differentiate between disease classes. CNN models exhibited the lowest AUC values, with ResNet50 (0.48–0.56) and InceptionV3 (0.51–0.54), confirming their limited capability in real-world disease classification.

Similarly, the MCC scores in Fig.~\ref{fig:mcc_scores} provide further insights into model reliability. ViT-B-16 achieved the highest MCC score (0.589), followed by CLIP-ViT-B-16 (0.491), highlighting their strong correlation with correct predictions. CNN models had the weakest correlation, with ResNet50 (0.089) and InceptionV3 (0.061), further demonstrating their limitations in capturing complex disease patterns.

\subsection{CLIP Interpretability: Correct vs. Incorrect Predictions}
This case study highlights one of the main advantages of CLIP interpretability over conventional image-only classification models: the integration of natural language descriptions as an additional diagnostic cue.

In the misclassified case, although the model prediction was incorrect (healthy instead of early blight), the textual description provided a clear statement of what the model perceived in the image. A farmer or agricultural expert, upon reading the description, could recognize the mismatch with the actual condition of the plant in the field. This offers a practical confirmation mechanism for the diagnostic outcome, helping users detect unreliable predictions.

In contrast, the correctly classified case shows the added value of CLIP’s interpretability, where the description explicitly highlighted the presence of early blight symptoms (brown circular lesions on aging foliage). Such explicit symptom descriptions make the prediction more trustworthy and informative compared to a label alone.

Therefore, beyond classification accuracy, the interpretability through natural language allows farmers to validate or question the model’s diagnosis, and even adjust their data collection practices (e.g., capturing closer or clearer images when symptoms are not well represented). This improves both the usability and robustness of AI-based plant disease detection systems.

\subsection{The Challenge of Bridging the Gap Between Controlled and Real-World Data}

A major challenge in plant disease classification lies in adapting models trained on controlled datasets (e.g., PlantVillage) to real-world, farmer-collected images. Convolutional Neural Networks (CNNs), despite achieving strong results on benchmark datasets, often fail to generalize due to their reliance on local feature extraction and sensitivity to background noise. In contrast, vision transformers demonstrate greater robustness by capturing global contextual relationships across an image, thereby improving performance under varied field conditions.

CLIP zero-shot models extend this advantage by bridging the gap between controlled and real-world domains without requiring task-specific training. Instead of relying solely on labeled datasets, CLIP leverages natural language descriptions to guide classification, embedding both images and text in a shared semantic space. This dual-modality framework enables rapid adaptation to new disease categories, provided the textual prompts are sufficiently descriptive.

Beyond classification performance, CLIP introduces a critical element of interpretability. Unlike conventional deep learning models that output opaque class labels, CLIP grounds its predictions in human-readable textual descriptions of symptoms. This property allows end-users to validate the plausibility of a diagnosis directly against their own observations, enhancing trust and supporting informed decision-making in agricultural practice. Such interpretability is particularly valuable in real-world farming environments, where variability in image quality and conditions can otherwise limit the reliability of automated tools.


\section{Conclusion} \label{sec5}
In this study, we evaluated the real-world performance of CNN-based, Transformer-based, and CLIP-based zero-shot models for plant disease classification. Although CNNs have been widely used in earlier research, our findings demonstrate their limitations in complex environments, particularly with field images captured by farmers under natural conditions. In contrast, transformer models and CLIP zero-shot architectures, which classify images based on textual descriptions without training, exhibited superior generalization and flexibility, making them more suitable for practical agricultural applications. This work highlights the limitations of training solely on idealized datasets like PlantVillage and underscores the importance of evaluating models under realistic conditions. By testing across varied complexities, we contribute to bridging the gap between controlled experiments and real-world deployment. For future research, we plan to explore lightweight transformer and zero-shot models optimized for real-time, in-field use. Additionally, we aim to expand the Farmy dataset by collecting a more diverse range of real-world images across various crops and disease stages, enabling broader and more practical benchmarking. These efforts will contribute to building scalable, accessible plant disease detection tools that support better crop health and food security.


\section*{Data availability statement}
\begin{itemize}

\item PlantVillage\cite{ref1} 

\url{https://github.com/spMohanty/PlantVillage-Dataset/tree/master/raw}
\item Farmy Dataset

\url{https://zenodo.org/records/10719742}
\item Africa Dataset\cite{ref3}

\url{https://zenodo.org/records/10719742} 
\item Peru Dataset\cite{ref3}

\url{https://zenodo.org/records/10719742}
\item Internet Dataset

\url{https://zenodo.org/records/10719742}

\end{itemize}

\section*{Declarations}

\begin{itemize}

\item Conflict of interest:

The authors declare that they have no conflict of interest
\end{itemize}


\end{document}